\documentclass[conference]{IEEEtran}
\IEEEoverridecommandlockouts
\usepackage{cite}
\usepackage{multirow}
\usepackage{amsmath,amssymb,amsfonts}
\usepackage{algorithmic}
\usepackage{graphicx}
\usepackage{url}
\usepackage{textcomp}
\usepackage{xcolor}
\def\BibTeX{{\rm B\kern-.05em{\sc i\kern-.025em b}\kern-.08em
    T\kern-.1667em\lower.7ex\hbox{E}\kern-.125emX}}

\newcommand{\myposspace}{\vspace*{0.15cm}}
\newcommand{\eg}{\textit{e.g.,}}

\begin{document}

\title{BRep Boundary and Junction Detection for CAD Reverse Engineering\\
\thanks{Corresponding Author: Sk Aziz Ali (www.skazizali.com)}
}

\author{
\IEEEauthorblockN{Sk Aziz Ali}
\IEEEauthorblockA{\textit{German Research Center for}\\ \textit{Artificial Intelligence (DFKI GmbH)}
\\sk\_aziz.ali@dfki.de}
\and
\IEEEauthorblockN{Mohammad Sadil Khan}
\IEEEauthorblockA{\textit{RPTU Kaiserslautern-Landau}\\Mohammad\_Sadil.Khan@dfki.uni-kl.de}
\and
\IEEEauthorblockN{Didier Stricker}
\IEEEauthorblockA{\textit{German Research Center for}\\ \textit{Artificial Intelligence (DFKI GmbH)}
\\didier.stricker@dfki.de}
}

\maketitle

\begin{abstract}
In machining process, 3D reverse engineering of the mechanical system is an integral, highly important, and yet time consuming step to obtain parametric CAD models from 3D scans. Therefore, deep learning-based Scan-to-CAD modeling can offer designers enormous editability to quickly modify CAD model, being able to parse all its structural compositions and design steps. In this paper, we propose a supervised boundary representation (BRep) detection network BRepDetNet from 3D scans of CC3D and ABC dataset. We have carefully annotated the $\sim$50K and $\sim$45K scans of both the datasets with appropriate topological relations (\textit{e.g.,} next, mate, previous) between the geometrical primitives (\textit{i.e.,} boundaries, junctions, loops, faces) of their BRep data structures. The proposed solution decomposes the Scan-to-CAD problem in Scan-to-BRep ensuring the right step towards feature-based modeling, and therefore, leveraging other existing BRep-to-CAD modeling methods. Our proposed Scan-to-BRep neural network learns to detect BRep boundaries and junctions by minimizing focal-loss and non-maximal suppression (NMS) during training time. Experimental results show that our BRepDetNet with NMS-Loss achieves impressive results.

\end{abstract}

\vspace{0.2cm}
\begin{IEEEkeywords}
BRep, Boundary Detection, Junction Detection, Scan-to-CAD, Reverse Engineering, NMS.
\end{IEEEkeywords}

\section{Introduction}
\label{sec:intro}
%
General purpose computer-aided design (CAD) modeling, using many modern APIs \cite{solidworks, autocad, Fusion360API, ParaSolid}, greatly assists in the agile process of Computer-Aided Manufacturing (CAM) and commercial applications~\cite{GeomagicDX} at industrial scale. Such a process requires skilled engineers and designers to aid in the creation, modification, analysis, or optimization of a design~\cite{sarcar2008computer}. The same expertise is needed when the physical object of the design is available. In this context, there has been a huge recent interest in CAD \textit{reverse engineering}~\cite{dubravvcik2012application}, which was made possible thanks to the availability of highly accurate 3D scanners\footnote{\url{https://www.artec3d.com/}}. 

\textit{How important is fast track \lq3D Reverse Engineering (RE)\rq~of mechanical system?} While the machining process starts with ready-made CAD models and ends with component manufacturing, RE starts with 3D scanning of either damaged parts, archived components, or any spare parts and ends with deducing its parametric CAD model. However, the purpose of RE is not limited to scanning a physical object to obtain a reliable 3D scan, and then concluding the CAD model. Being able to deduce -- (i) the intermediate designing steps, (ii) the history of modifications that are involved in parametric modeling of geometric primitives, (iii) designer's intent to order the CAD primitives (\eg, plane, cylinders, cones) along with different CAD operation steps (\eg, chamfer, extrusion, revolution, fillet) are the main stages of RE. Therefore, RE can fix the damaged parts of any mechanical system through fast-track prototyping, mass manufacturing, quality analysis, and other downstream application tasks for analyzing structural compositions of real objects. 


%

\textit{Why Boundary Representation (BRep) of CAD is ideal for deep learning-based RE?} 3D scans are unstructured representations and are 
different from CAD models, which are generally stored in their STandard for the Exchange of Product (STEP)~\cite{BHANDARKAR20003} file format and represented as Boundary-Representations (BReps)~\cite{zonegraph}. Consequently, the first step towards successful CAD reverse engineering 
is to learn/infer BRep~\cite{zonegraph} representation from its corresponding 3D scan, namely \textit{Scan-to-BRep}. This step is considered as the gateway for numerous downstream CAD modeling applications like -- grouping and segmentation of BRep faces~\cite{lambourne2021brepnet,CADOPsNetDV22,uvnet}, deducing machining features~\cite{HierachicalCADNet22}, retrieving CAD sketches~\cite{CADOPsNetDV22}, predicting mating components for assembling multiple CAD models~\cite{SBGCN21}.

Our proposed method brings the problem of 3D scan to BRep parsing closer to real-world settings. In particular, we annotate $\sim$$50$K real 3D scans of CC3D \cite{cc3d} and $\sim$$45$K high-resolution meshes of ABC \cite{abc_dataset} datasets with BRep edge, face, and their topological information. In particular, a per-vertex annotation is considered in terms of BRep boundaries and junctions. To understand the BRep chain complex, in other words, the adjacency systems between BRep \textit{body, face, boundary, junction, co-edge/boundary loop, shell, fin, and region} entities, one must parse it into different \lq\textit{closed and connected loops of co-edges} (CCLC)\rq. Therefore, once the boundary loops and junction points are detected in a given scan, the next steps like segmenting and grouping them into their respective BRep entities, fitting parametric curves and surfaces on the segmented boundaries and faces~\cite{guo2022complexgen, wang2020pie}, inferring the CAD construction steps and operation types involved at each step~\cite{CADOPsNetDV22}, and retrieval of base sketches~\cite{point2cyl} follow on. Parsing BRep from 3D scans can also be seen as geometric computation with its \textit{chain complexes}~\cite{furiani2017geometric}. Once the BRep representation is learned, we gain complete control BRep chain complex (refer to \cite{guo2022complexgen} for the definition). 

\myposspace
\noindent\textbf{Contributions.} \textbf{(1)} We contribute in preparing annotations (ensuring through quality measurements) for deriving BRep chain complex on industrial-level 3D scans from CC3D \cite{cc3d} and ABC \cite{abc_dataset} datasets. \textbf{(2)} Besides, we propose a new network for \textit{boudary and junction} points detection from a 3D scan. \textbf{(3)} For this task, the existing state-of-the-art methods~\cite{wang2020pie, tan2022coarse} rely upon non-maximal suppression (NMS) as a post-prediction step to heuristically prune large number of incorrect predictions. In contrast, our detection network includes a novel way to marginalize false positives/negatives directly during training time. This results in impressive boundary and junction detection recall gains compared to other methods. \textbf{(4)} Source code is available here {\color{pink}\url{https://github.com/saali14/Scan-to-BRep}}.

\section{BRep Data Annotation Features}
\label{sec:Dataset}
\begin{figure}[!ht]
\begin{center}
   \includegraphics[width=0.99\linewidth]{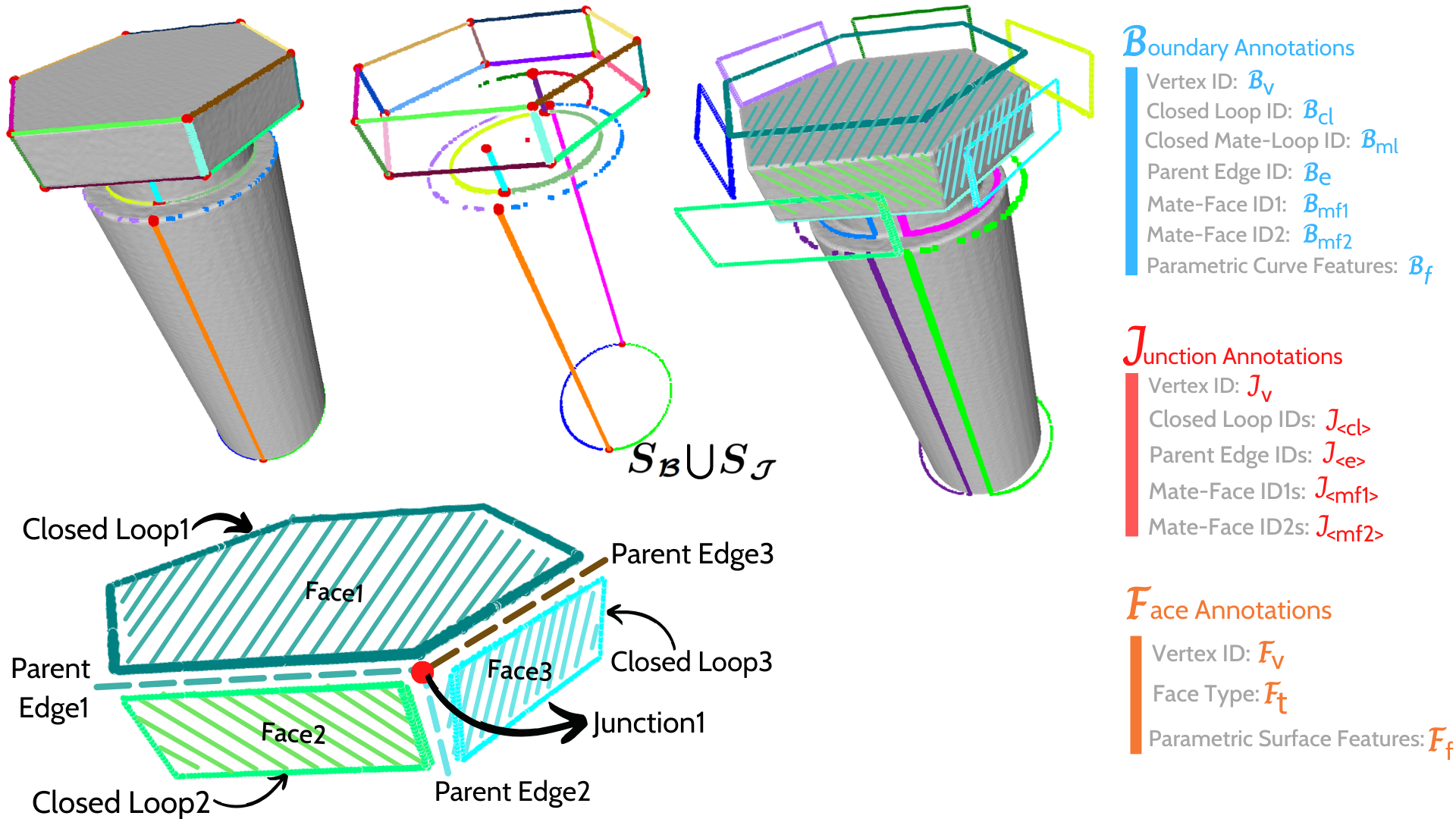}
\end{center}
\caption{The \textit{left} figures relate a 3D scan with its corresponding BRep topology (color-coded by unique edge and different closed-loop ids). A list of BRep annotations per scan point (on the \textit{right}) covers all attributes under CCLC.
}
\label{fig:AnnotationDetails}
\end{figure}
Formally, a BRep topology $\boldsymbol{\mathcal{T}}=(\boldsymbol{\mathcal{B}}, \boldsymbol{\mathcal{J}}, \boldsymbol{\mathcal{F}}, \boldsymbol{\mathcal{A}})$ is defined by sets of faces $\boldsymbol{\mathcal{F}}$, edges $\boldsymbol{\mathcal{E}}$, boundaries $\boldsymbol{\mathcal{B}}$, junctions $\boldsymbol{\mathcal{J}}$, and an adjacency system $\boldsymbol{\mathcal{A}}=(\boldsymbol{\mathcal{N}}, \boldsymbol{\mathcal{P}}, \boldsymbol{\mathcal{M}}) \in \left\lbrace0,1\right\rbrace^{{\small|\boldsymbol{\mathcal{B}}|\times|\boldsymbol{\mathcal{B}}|}}$
that ensures directed topological walk along the boundaries to traverse from current edge/face/junction to its \textit{next} $\boldsymbol{\mathcal{N}}$, \textit{previous} $\boldsymbol{\mathcal{P}}$, and \textit{mating} $\boldsymbol{\mathcal{M}}$ counterparts is possible. We redirect readers to ~\cite{lambourne2021brepnet} and the Figure \ref{fig:AnnotationDetails} to understand how different edges form a boundary, how boundaries form closed loops, and how loops form a closed CAD manifold. 
Each of the BRep topological entities are parametric -- \eg~edges are either \textit{spline, arc, lines, ellipse} or other types of curves, whereas faces are \textit{planar, B-Spline, parabolic} or other types of surfaces -- with unique ids based on forward/reverse order of the topological walk along the $\boldsymbol{\mathcal{B}}$.   
The figure~\ref{fig:AnnotationDetails} illustrates how a 3D scan $\boldsymbol{S}$ and the aforementioned BRep topology is related. For instance, the overlaid scan points $\boldsymbol{S}_{{\small\boldsymbol{\mathcal{B}}}}$ and $\boldsymbol{S}_{{\small\boldsymbol{\mathcal{J}}}}$, that are annotated as $\boldsymbol{\mathcal{B}}$ and $\boldsymbol{\mathcal{J}}$, are decomposed into closed-loops sharing sharing boundaries. We do not clutter the figure by showing rest of the scan points $\boldsymbol{S}_{{\small\boldsymbol{\mathcal{F}}}}$ that are annotated different face Ids from $\boldsymbol{\mathcal{F}}$. The list of attributes in figure~\ref{fig:AnnotationDetails} are stored for each scan point belonging to $\boldsymbol{\mathcal{B}}$ / $\boldsymbol{\mathcal{J}}$ / $\boldsymbol{\mathcal{F}}$.
Among the existing datsets~\cite{abc_dataset,cc3d,fusion360} that are dedicated to feature-based CAD modeling/reconstruction purposes, typical shortfalls are -- \textit{(i)} unavailability of 3D scans (often substituted by CAD models), \textit{(ii)} sharp-edge labels. Note that those boundaries which has with small acute angle between its adjacent BRep faces. The main difficulty of detecting boundary/junction points is that they can be smooth and seamless (see the vertical rule along the cylindrical barrel in Figure \ref{fig:AnnotationDetails}). \textit{(iii)} CAD construction step labels per BRep entity (mainly the faces), and \textit{(iv)} the CAD operations (\eg extrusion, fillet, cut, chamfer) involved in those construction steps, and \textit{(v)} proper segmentation labels of scan points \hbox{w.r.t} their respective BRep entities (as well as their parameters). 
The previous variants of CC3D dataset~\cite{cc3d, CADOPsNetDV22} provide annotations mentioned in \textit{(i) -- (iv)}. We have complemented annotations in \textit{(v)}.
%
%
%
%
We encourage the community to use our annotations in the future.


\subsection{Annotation Quality Assessment}\label{subsec:DatasetQualityAssesment} 
\begin{figure}[!ht]
\begin{center}
   \includegraphics[width=1.0\linewidth]{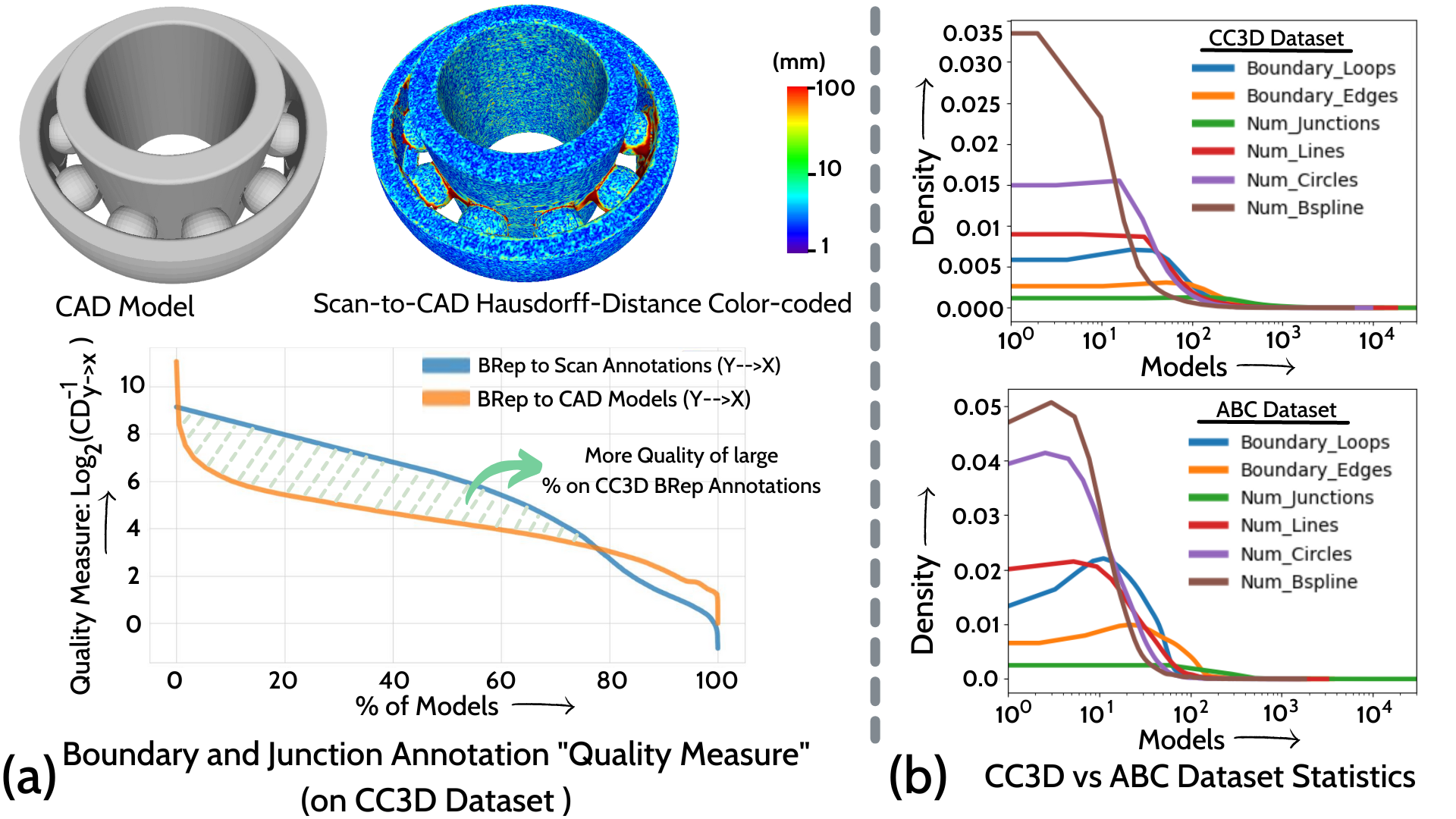}
\end{center}
\vspace{-0.7cm}
\caption{
}
\label{fig:DatasetStat_Quality}
\end{figure}
We perform a crucial annotation quality assessment over all samples of \textit{CC3D} dataset. For this, we compute inverse of uni-directional and density-aware Chamfer-distance~\cite{DCDWu2021} from parametric BRep Boundary $\boldsymbol{S}_{{\small\boldsymbol{\mathcal{B}}}}$ to 3D real scan $\boldsymbol{S}$ in logarithmic scale as:
\begin{equation}
\log_2\left(CD^{-1}\right) =  
\log_2
\left(\frac{1}{|\boldsymbol{S}_{{\small\boldsymbol{\mathcal{B}}}}|} \sum_{\boldsymbol{s}_i\in\boldsymbol{S}_{{\small\boldsymbol{\mathcal{B}}}}} \min_{\boldsymbol{p}\in\boldsymbol{S}}\left( 1- e^{\lVert \boldsymbol{s}_i -\boldsymbol{p}\rVert_2}\right)\right)^{-1},
\end{equation}
and compare the same measurement when computed from the parametric BRep Boundary $\boldsymbol{S}_{{\small\boldsymbol{\mathcal{B}}}}$ to the parametric CAD model of the scan. 
The figure~\ref{fig:DatasetStat_Quality}-(a) illustrates a larger percentage of models has significantly higher quality scores when annotations are compared against real scans. This re-affirms that our scan-to-BRep annotations are closer to the real-world setting. It is common that real scans are partial and noisy than their CAD versions which are regular/smooth and complete (see the color-coded Hausdorff distances from scan-to-CAD in figure ~\ref{fig:DatasetStat_Quality}-(a)). Therefore, it is necessary to not have annotations that are overly-smoothed like parametric BRep entities. Besides, the figure~\ref{fig:DatasetStat_Quality}-(b) also shows a comprehensive statistics on both \textit{CC3D} and \textit{ABC} models in terms of overall model complexities. For instance, it is clear that area covered under density plots for different sets parametric curves per model is several times more in \textit{CC3D} than \textit{ABC}.

\begin{table*}[]
\resizebox{\linewidth}{!}{\begin{tabular}{lcccc|cccc}
\hline
\multicolumn{1}{c}{Model} & \multicolumn{4}{c|}{ABC}                                          & \multicolumn{4}{c}{CC3D}                                          \\
\multicolumn{1}{c}{}      & \multicolumn{2}{c}{Boundary}    & \multicolumn{2}{c|}{Junction}   & \multicolumn{2}{c}{Boundary}    & \multicolumn{2}{c}{Junction}    \\
\multicolumn{1}{c}{}      & Recall         & Precision      & Recall         & Precision      & Recall         & Precision      & Recall         & Precision      \\ \hline
ComplexGen \cite{guo2022complexgen}               & 0.104          & 0.334          & 0.045          & 0.232          & 0.056          & 0.461          & 0.013          & 0.097          \\
PieNet  \cite{wang2020pie}                  & 0.338          & 0.391          & -              & -              & 0.430          & 0.529          & -              & -              \\
BRepDetNet                & \textbf{0.615} & \textbf{0.523} & \textbf{0.361} & \textbf{0.273} & \textbf{0.747} & \textbf{0.631} & \textbf{0.427} & \textbf{0.364} \\ \hline
\end{tabular}}
\vspace{0.15cm}
\caption{Recall and Precision for the boundary and junction detection on ABC\cite{abc_dataset} (left) and CC3D\cite{cc3d} (right) dataset. All the models are trained on ABC\cite{abc_dataset} dataset.}
\label{tab:table_baseline}
\end{table*}

\begin{table*}[]
\resizebox{\linewidth}{!}{\begin{tabular}{lllll|llll}
\hline
\multicolumn{1}{c}{Model} & \multicolumn{4}{c|}{Model Trained on ABC dataset}                                                                        & \multicolumn{4}{c}{Model Trained on CC3D dataset}                                                                       \\
\multicolumn{1}{c}{}      & \multicolumn{2}{c}{ABC}                                    & \multicolumn{2}{c|}{CC3D}                                   & \multicolumn{2}{c}{ABC}                                    & \multicolumn{2}{c}{CC3D}                                   \\
                          & \multicolumn{1}{c}{Recall} & \multicolumn{1}{c}{Precision} & \multicolumn{1}{c}{Recall} & \multicolumn{1}{c|}{Precision} & \multicolumn{1}{c}{Recall} & \multicolumn{1}{c}{Precision} & \multicolumn{1}{c}{Recall} & \multicolumn{1}{c}{Precision} \\ \hline
Ours w/o NMS Loss   &                            0.454&                               0.237&                            0.622&                                0.383&                            \textbf{0.502}&                               0.380&                            0.692&                               0.646\\
Ours                      &                            \textbf{0.615}&                               \textbf{0.523}&                            \textbf{0.747}&                                \textbf{0.631}&                            0.370&                               \textbf{0.510}&                            \textbf{0.697}&                               \textbf{0.652}\\ \hline
\end{tabular}}
\vspace{0.15cm}
\caption{Quantitative results for BRepDetNet with or without NMS Loss. Recall and Precision for boundary detection on ABC \cite{abc_dataset} and CC3D \cite{cc3d} dataset are reported. We also showcase cross-dataset generalization ability of our model.}
\label{tab:table_2}
\end{table*}

\section{B-Rep Boundary and Junction Detection}
\label{sec:Boundary_Detection}
\begin{figure}[!ht]
\begin{center}
\includegraphics[width=0.99\linewidth]{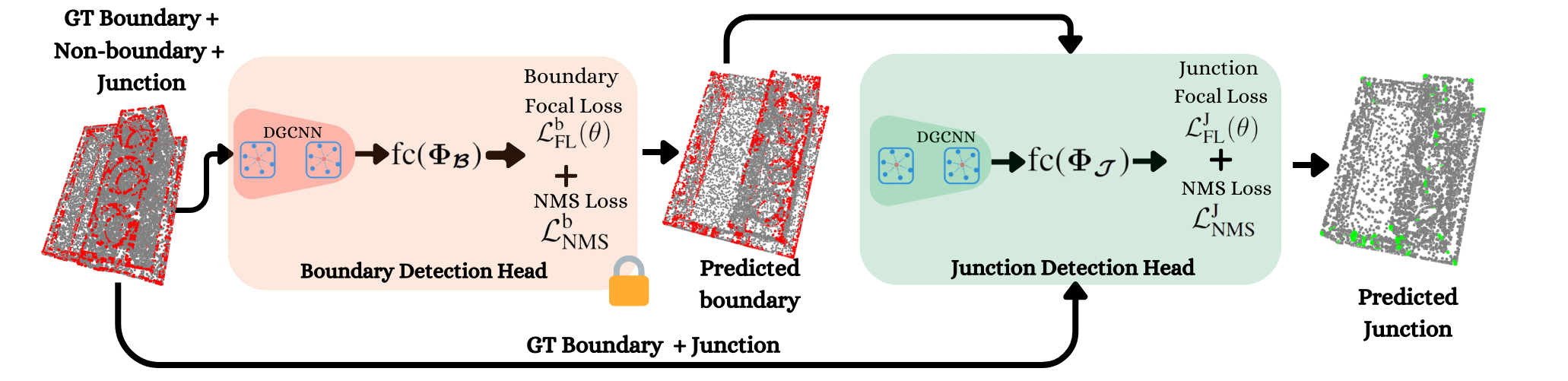}
\end{center}
\vspace{-0.45cm}
\caption{Our network architecture.
}
\label{fig:BRepDetNet_Arch}
\end{figure}
\subsection{Network Architecture of BRepDetNet}
Figure \ref{fig:BRepDetNet_Arch} illustrates that our neural network is comprised of separate boundary detection and junction detection heads. Both network heads use DGCNN~\cite{dgcnn} as point feature encoder that maps $\mathcal{D}:\boldsymbol{S}\mapsto \mathbf{\Phi} \in \mathbb{R}^{N\times128}$ resulting into 128 dimensional point-level deep feature vectors. The two separate DGCNN encoders output boundary and junction embedding vectors $\mathbf{\Phi}_{{\small\boldsymbol{\mathcal{B}}}}$ and $\mathbf{\Phi}_{{\small\boldsymbol{\mathcal{J}}}}$ respectively. Next, we apply fully-connected layer to resize fc($\mathbf{\Phi}_{{\small\boldsymbol{\mathcal{B}}}}$) and fc($\mathbf{\Phi}_{{\small\boldsymbol{\mathcal{J}}}}$) with final dimensions in $\mathbf{R}^{N\times1}$.

\subsection{BRep Boundary Detection}\label{subsec:BRep_Boundary_Det}
To formalize the problem of classifying each point as either a boundary point or an interior point of a 3D scan $\boldsymbol{S}\in\mathbb{R}^{N\times3}$, we learn  a function $f_{\theta}:\mathbb{R}^{3}\rightarrow\{0,1\}$ that maps each point $\mathbf{p}_i$ to a label $y_i$. Here \( y_i = 0 \) indicates an interior point, and \( y_i = 1 \) indicates a boundary point.
The function \( f_{\theta} \) is parameterized by a neural network weights \( \theta \). We use
focal loss (FL) \cite{FocalLoss} as an extension of the binary cross-entropy loss and is designed to address class imbalance. It adds a modulating factor \( (1 - p_t)^{\gamma} \) to the cross-entropy loss, where \( p_t \) is the true class probability and \( \gamma \) is the focusing parameter as:
\begin{equation}
\begin{split}
\mathcal{L}_{\text{FL}}^{\text{b}}(\theta) = 
& -\frac{1}{N} \sum_{i=1}^{N}  (1 - f_{\theta}(\mathbf{p}_i))^{\gamma} \cdot y_i \log(f_{\theta}(\mathbf{p}_i)) \\ 
& + (1 - (1 - f_{\theta}(\mathbf{p}_i))^{\gamma}) \cdot (1-y_i) \log(1 - f_{\theta}(\mathbf{p}_i)).    
\end{split}
\end{equation}

Non-Maximal Suppression (NMS)~\cite{neubeck2006efficient} technique is primarily used in object detection tasks to eliminate multiple bounding boxes that are close to each other and keep only the most probable one. We have introduced NMS loss for boundary detection \( \mathcal{L}^{b}_{\text{NMS}}(\theta) \) during the training time as a differentiable approximation to the actual NMS operation:
\begin{equation}\label{eq:boundary_NMS}
\mathcal{L}_{\text{NMS}}^{\text{b}} = -\frac{1}{N}\sum_{i=1}^{N}\left(\alpha t_i\log(p_i) + (1 - \alpha)(1 - t_i)\log(1 - p_i)\right)  
\end{equation}

$L_{\text{NMS}}^{\text{b}}(\theta)$ represents the Non-Maximal Suppression Classification loss for a particular prediction $p_i = f_{\theta}(\mathbf{p}_i)$ and target label $t_i$.
$N$ is the total number of predictions, where
$p_i$ is the predicted probability of a point being at the boundary of $\boldsymbol{S}$ and $\alpha$ is a hyperparameter that controls the balance between the two terms in the loss. It's typically set between 0.25 and 0.75. $L_{\text{NMS}}^{\text{b}}(\theta)$ penalizes both false positives and negatives. This encourages the model to produce more accurate boundary scores. Therefore our total loss for boundary detection is 
\begin{equation}\label{eq:boundary_total}
\mathcal{L}_{\text{total}}^{\text{b}} =  \mathcal{L}_{\text{FL}}^{\text{b}} + \mathcal{L}_{\text{NMS}}^{\text{b}}
\end{equation}

\subsection{B-Rep Junction Detection}
\label{sec:Boundary_Detection}
Detecting the B-Rep junction points is considered as a subordinate task after boundary detection. Let \( \mathcal{B} \) be the set of points detected as boundary points (note, this includes \textit{false positives} but not the \textit{false negative} points). A second neural network \( g_{\phi}(\mathbf{x}) \) is trained to classify these as either corner points or not. The loss function for this second task is also the binary cross-entropy but restricted to the boundary points:
\begin{equation}
\begin{split}
\mathcal{L}_{\text{FL}}^{\text{J}}(\theta) = 
& -\frac{1}{|\mathcal{B}|}  \sum_{\mathbf{x}_i \in \mathcal{B}}   (1 - g_{\phi}(\mathbf{x}_i))^{\gamma} \cdot z_i \log(g_{\phi}(\mathbf{x}_i)) \\ 
& + (1 - (1 - g_{\phi}(\mathbf{x}_i))^{\gamma}) \cdot (1-z_i) \log(1 - g_{\phi}(\mathbf{x}_i)).    
\end{split}
\end{equation}
where $z_i$ denotes label values for $\mathbf{x}_i$ being a junction or not. Similarly, the NMS loss $\mathcal{L}_{\text{NMS}}^{\text{J}}$ and total loss $\mathcal{L}_{\text{total}}^{\text{J}}$ for junction detection are equivalent to the Eq.~\eqref{eq:boundary_NMS} and \eqref{eq:boundary_total} respectively.

\section{Experimental Results}\label{sec:Experimental_Results}
We have used 10K randomly sampled scan points as input to out boundary detection head and 4K randomly sampled points from the detected boundaries to the junction detection head. Suppose, during inference time, if the resulting number of boundary points is below the required number of 4K input points, they are randomly up-sampled with duplicated matches. We train our network with batch size 10 and learning rate 0.001 on a 48GB NVIDIA GPU. 

\subsection{Baseline Models}
We evaluate our proposed approach on ComplexGen\cite{guo2022complexgen} and PieNet\cite{wang2020pie}. Both the baseline models and our proposed BRepDetNet are trained on ABC\cite{abc_dataset} dataset and tested on ABC\cite{abc_dataset} and CC3D \cite{cc3d} dataset for a consistent comparison. For PieNet, we only show the results for boundary prediction due to its unstable training and crashing during inference for junction detection. 

\subsection{Comparison with The Baseline Models}\label{subsec:BaselineComparison}
\begin{figure*}
\centering
\includegraphics[width=0.99\textwidth]{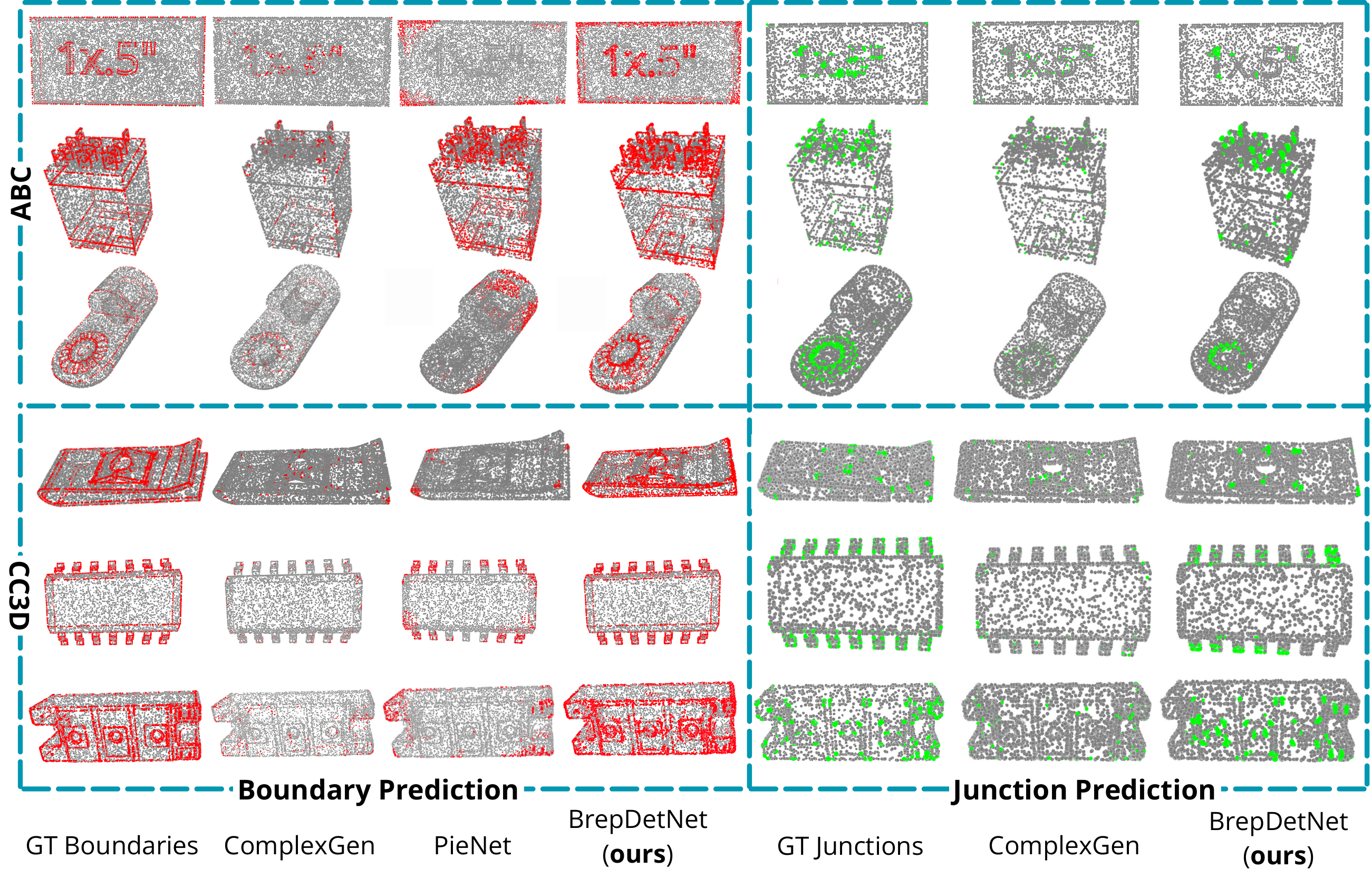}
\caption{Visual results for boundary (left) and junction (right) prediction on ABC~\cite{abc_dataset} (top) and CC3D~\cite{cc3d} (bottom) dataset. The red points are the boundary points and the green points are the junction points. All the models have been trained on ABC~\cite{abc_dataset} dataset.} \label{fig:table_1_figure}
\end{figure*}
\begin{figure*}[!ht]
\centering
\includegraphics[width=0.99\textwidth]{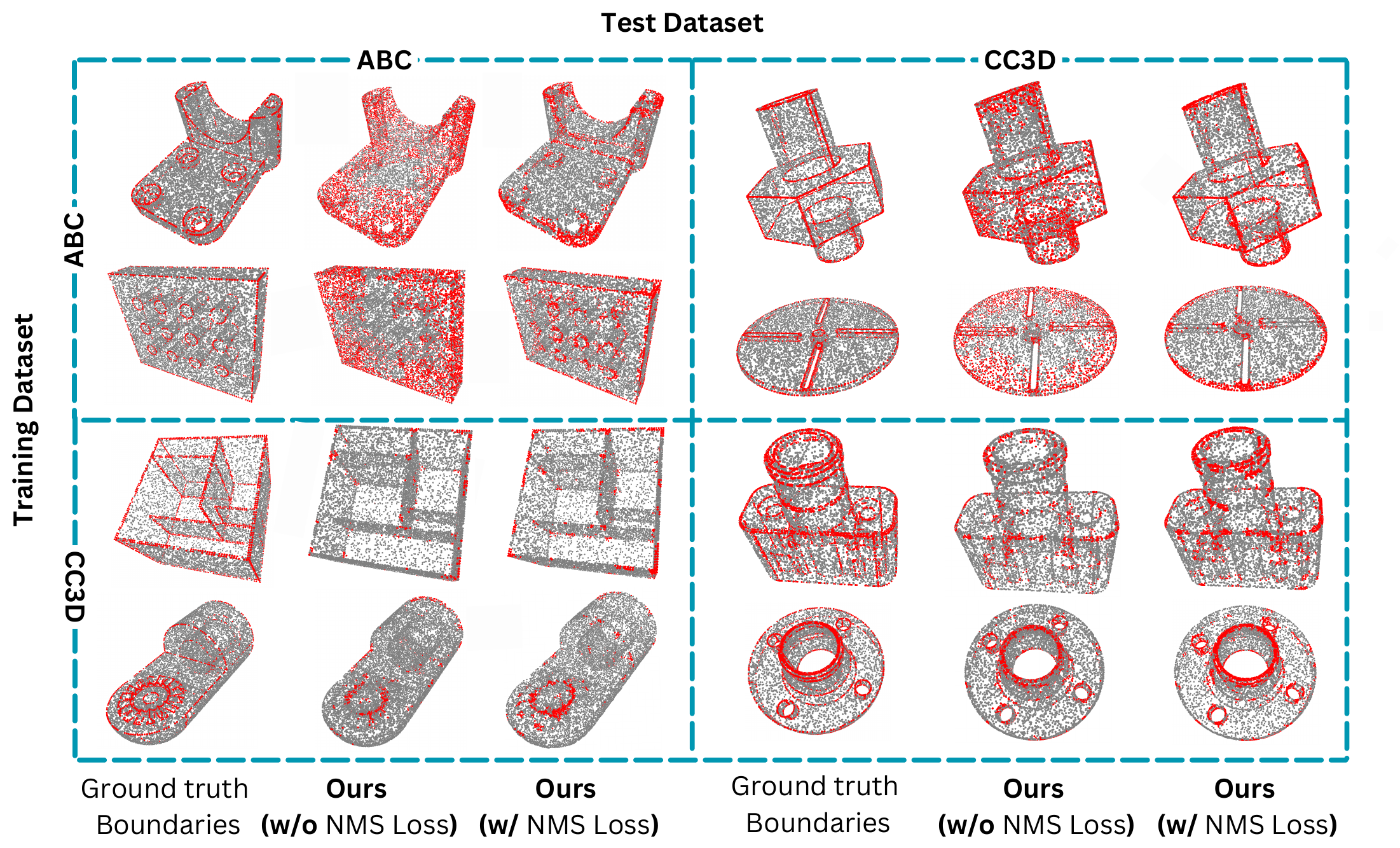}
\caption{Visual results for boundary detection for BRepDetNet with and without NMS Loss. The red points are the boundary points. We observe that using NMS loss significantly reduces the number of false positives leading to higher precision in boundary detection.} \label{fig:table_2_figure}
\end{figure*}
Table \ref{tab:table_baseline} provides the quantitative results on boundary and junction prediction tasks for ABC\cite{abc_dataset} and CC3D\cite{cc3d} dataset. We measure \textit{recall} and \textit{precision} metrics as 
\[
\text{Recall} = \frac{\text{True Positives}}{\text{True Positives + False Negatives}}
\]
\[
\text{Precision} = \frac{\text{True Positives}}{\text{True Positives + False Positives}}
\] 
to evaluate the performance of the baseline methods.
As shown in Table \ref{tab:table_baseline}, our method outperforms the baselines in recall and precision for both boundary and junction detection. Our method shows significantly higher accuracy in terms of \textit{precision-recall} combined measure. For instance, ComplexGen method detects fewer boundary points and miss to detect a large amount. Although, it does not detect false positives/negatives as compared to PieNet. In contrast to boundary prediction, detecting junction points is more difficult as they are often less than $\sim1\%$ of the total boundary points. As in Table \ref{tab:table_baseline}, BRepDetNet shows significant improvement in recall compared to the state-of-the-art methods for junction detection tasks. We also observe that despite being trained on a clean and noise free meshes of ABC\cite{abc_dataset} dataset, our model generalizes well on real 3D scans. The result surpasses that of baseline models in cross-dataset experiments on CC3D\cite{cc3d}. 

Figure \ref{fig:table_1_figure} shows some qualitative results between the baseline models and BrepDetNet. As shown, our model predicts more accurate boundary points (red points) compared to other methods. Furthermore, it can be seen that most of the false positive predictions for BRepDetNet are close to the ground-truth boundary points. Thanks to the NMS loss, furthest non-boundary points are not misclassified like PieNet\cite{wang2020pie}.


\subsection{Ablation Study}\label{subsec:Ablation_Study}
%
In this section, we showcase the effectiveness of NMS loss and . Table \ref{tab:table_2} reports the precision and recall for the boundary prediction using BRepDetNet with and without NMS loss. In Figure ~\ref{fig:table_2_figure}, some visual results are provided. To evaluate the performance, we trained both models on ABC~\cite{abc_dataset} and tested on both ABC~\cite{abc_dataset} and CC3D~\cite{cc3d} datasets. We use the same strategy with CC3D~\cite{cc3d} dataset. This results in four different experimental scenarios as shown in Table~\ref{tab:table_2}. We observe that using NMS Loss significantly improves the precision in all four scenarios. This can be explained by the reduction of misclassified boundary points. As demonstrated in Figure~\ref{fig:table_2_figure}, NMS loss leads to a lower number of false positives.

\section{Limitation and Conclusion}
In this paper, we propose BRepDetNet, a supervised boundary and junction detection network as the Scan-to-BRep step for 3D reverse engineering. We offer both topological and geometrical annotations that relates 3D real scans with its BRep chain complex. Our proposed approach outperforms the baseline models on both ABC\cite{abc_dataset} and CC3D\cite{cc3d} datasets. As part of future extension, we plan to make our network end-to-end trainable and resilient even when the annotation quality is poor. We also notice some degree of mis-classification of boundary points directly affect the subsequent junction prediction. Therefore, improving a joint learning of boundary and junction detection tasks, followed by segmentation, and grouping of BRep entities will also be part of our future work.  
\section{Acknowledgements}
This work was partially funded by the EU Horizon Europe
Framework Program under grant agreement 101058236 (HumanTech). I am grateful to Prof. Djamila Aouada and Dr. Anis Kacem (Snt, University of Luxembourg) for their valuable inputs in understanding Scan-to-BRep paradigm.

\newpage
{\small
\bibliographystyle{ieee_fullname}
\bibliography{egbib}

\begin{thebibliography}{10}\itemsep=-1pt

\bibitem{GeomagicDX}
3D~Systems~Inc. 2021.
\newblock {geomagic Design X}.
\newblock \url{https://www.3dsystems.com/software/geomagic-design-x}, 2021.

\bibitem{autocad}
AutoCAD.
\newblock {AutoCAD: 2D and 3D CAD}.
\newblock \url{https://www.autodesk.com/products/autocad/}.

\bibitem{Fusion360API}
Autodesk.
\newblock {Autodesk Fusion 360 API }.
\newblock \url{https://autodeskfusion360.github.io/}, 2014.

\bibitem{BHANDARKAR20003}
Mangesh~P. Bhandarkar and Rakesh Nagi.
\newblock Step-based feature extraction from step geometry for agile
  manufacturing.
\newblock {\em Computers in Industry}, 41(1):3--24, 2000.

\bibitem{cc3d}
K. Cherenkova, D. Aouada, and G. Gusev.
\newblock Pvdeconv: Point-voxel deconvolution for autoencoding cad construction
  in 3d.
\newblock In {\em International Conference on Image Processing (ICIP)}, 2020.

\bibitem{HierachicalCADNet22}
Andrew~R. Colligan, Trevor~T. Robinson, Declan~C. Nolan, Yang Hua, and Weijuan
  Cao.
\newblock Hierarchical cadnet: Learning from b-reps for machining feature
  recognition.
\newblock {\em Computer-Aided Design}, 147:103226, 2022.

\bibitem{dubravvcik2012application}
Michal D{\'u}brav{\v{c}}{\'\i}k and {\v{S}}tefan Kender.
\newblock Application of reverse engineering techniques in mechanics system
  services.
\newblock {\em Procedia Engineering}, 48:96--104, 2012.

\bibitem{CADOPsNetDV22}
Elona Dupont, Kseniya Cherenkova, Anis Kacem, Sk~Aziz Ali, Ilya Aryhannikov,
  Gleb Gusev, and Djamila Aouada.
\newblock Cadops-net: Jointly learning cad operation types and steps from
  boundary-representations.
\newblock In {\em 2022 International Conference on 3D Vision (3DV)}, 2022.

\bibitem{furiani2017geometric}
Francesco Furiani, Giulio Martella, and Alberto Paoluzzi.
\newblock Geometric computing with chain complexes: Design and features of a
  julia package.
\newblock {\em arXiv preprint arXiv:1710.07819}, 2017.

\bibitem{guo2022complexgen}
Haoxiang Guo, Shilin Liu, Hao Pan, Yang Liu, Xin Tong, and Baining Guo.
\newblock Complexgen: Cad reconstruction by b-rep chain complex generation.
\newblock {\em ACM Transactions on Graphics (TOG)}, 41(4):1--18, 2022.

\bibitem{uvnet}
Pradeep~Kumar Jayaraman, Aditya Sanghi, Joseph~G. Lambourne, Karl~D.D. Willis,
  Thomas Davies, Hooman Shayani, and Nigel Morris.
\newblock Uv-net: Learning from boundary representations.
\newblock In {\em CVPR}, 2021.

\bibitem{SBGCN21}
Benjamin~T. Jones, Dalton Hildreth, Duowen Chen, Ilya Baran, Vova Kim, and
  Adriana Schulz.
\newblock {SB-GCN:} structured {BREP} graph convolutional network for automatic
  mating of {CAD} assemblies.
\newblock {\em CoRR}, abs/2105.12238, 2021.

\bibitem{abc_dataset}
S. Koch, A. Matveev, Y. Jiang, F. Williams, A. Artemov, E. Burnaev, M. Alexa,
  D. Zorin, and D. Panozzo.
\newblock Abc: A big cad model dataset for geometric deep learning.
\newblock In {\em In Computer Vision and Pattern Recognition (CVPR)}, 2019.

\bibitem{lambourne2021brepnet}
Joseph~G. Lambourne, Karl~D.D. Willis, Pradeep~Kumar Jayaraman, Aditya Sanghi,
  Peter Meltzer, and Hooman Shayani.
\newblock Brepnet: A topological message passing system for solid models.
\newblock In {\em Proceedings of the IEEE/CVF Conference on Computer Vision and
  Pattern Recognition (CVPR)}, pages 12773--12782, June 2021.

\bibitem{FocalLoss}
T. Lin, P. Goyal, R. Girshick, K. He, and P. Dollar.
\newblock Focal loss for dense object detection.
\newblock In {\em 2017 IEEE International Conference on Computer Vision
  (ICCV)}, 2017.

\bibitem{neubeck2006efficient}
Alexander Neubeck and Luc Van~Gool.
\newblock Efficient non-maximum suppression.
\newblock In {\em International conference on pattern recognition (ICPR)},
  volume~3, pages 850--855. IEEE, 2006.

\bibitem{sarcar2008computer}
MMM Sarcar, K~Mallikarjuna Rao, and K~Lalit Narayan.
\newblock {\em Computer aided design and manufacturing}.
\newblock PHI Learning Pvt. Ltd., 2008.

\bibitem{ParaSolid}
Siemens.
\newblock {ParaSolid}.
\newblock \url{https://www.autodesk.com/products/autocad/}.

\bibitem{solidworks}
Solidworks.
\newblock {3D CAD Design Software}.
\newblock \url{https://www.solidworks.com/}.
\newblock Online: accessed 02-June-2022.

\bibitem{tan2022coarse}
Xuefeng Tan, Dejun Zhang, Long Tian, Yiqi Wu, and Yilin Chen.
\newblock Coarse-to-fine pipeline for 3d wireframe reconstruction from point
  cloud.
\newblock {\em Computers \& Graphics}, 106:288--298, 2022.

\bibitem{point2cyl}
Mikaela~Angelina Uy, Yen-Yu Chang, Minhyuk Sung, Purvi Goel, Joseph Lambourne,
  Tolga Birdal, and Leonidas Guibas.
\newblock Point2cyl: Reverse engineering 3d objects from point clouds to
  extrusion cylinders.
\newblock In {\em 2012 IEEE Conference on Computer Vision and Pattern
  Recognition (CVPR)}, 2022.

\bibitem{wang2020pie}
Xiaogang Wang, Yuelang Xu, Kai Xu, Andrea Tagliasacchi, Bin Zhou, Ali
  Mahdavi-Amiri, and Hao Zhang.
\newblock Pie-net: Parametric inference of point cloud edges.
\newblock {\em Advances in neural information processing systems},
  33:20167--20178, 2020.

\bibitem{dgcnn}
Y. Wang, Y. Sun, Z. Liu, S.~E. Sarma, M.~M. Bronstein, and Justin~M. Solomon.
\newblock Dynamic graph cnn for learning on point clouds.
\newblock {\em ACM Transactions on Graphics (TOG)}, 2019.

\bibitem{fusion360}
Karl D.~D. Willis, Yewen Pu, Jieliang Luo, Hang Chu, Tao Du, Joseph~G.
  Lambourne, Armando Solar-Lezama, and Wojciech Matusik.
\newblock Fusion 360 gallery: A dataset and environment for programmatic cad
  construction from human design sequences.
\newblock {\em ACM Trans. Graph.}, 40(4), jul 2021.

\bibitem{DCDWu2021}
T. Wu, L. Pan, J.~Zhang~T. WANG, Z. Liu, and D. Lin.
\newblock Density-aware chamfer distance as a comprehensive metric for point
  cloud completion.
\newblock In {\em In Advances in Neural Information Processing Systems
  (NeurIPS), 2021}, 2021.

\bibitem{zonegraph}
Xianghao Xu, Wenzhe Peng, Chin-Yi Cheng, Karl~DD Willis, and Daniel Ritchie.
\newblock Inferring cad modeling sequences using zone graphs.
\newblock In {\em Proceedings of the IEEE/CVF Conference on Computer Vision and
  Pattern Recognition}, pages 6062--6070, 2021.

\end{thebibliography}
}

\end{document}